\pdfoutput=1

\documentclass[11pt]{article}

\usepackage[]{EMNLP2022}

\usepackage{times}
\usepackage{latexsym}
\usepackage{multirow}
\usepackage{tabularx}
\usepackage{graphicx}
\usepackage{subfigure}
\usepackage{tabularx}
\usepackage{booktabs}
\usepackage{stfloats}
\usepackage{xspace}
\usepackage{amsmath, amssymb}
\newtheorem{theorem}{Theorem}
\usepackage{microtype}
\usepackage[T1]{fontenc}

\usepackage[utf8]{inputenc}
\usepackage{microtype}
\usepackage{inconsolata}

\newcommand{\modelname}{\textsc{Trace}\xspace}
\newcommand*{\dif}{\mathop{}\!\mathrm{d}}

\title{Recurrence Boosts Diversity! Revisiting Recurrent Latent Variable in Transformer-Based Variational AutoEncoder for Diverse Text Generation}

\author{
    Jinyi Hu$^{1,2,3}$,\,  Xiaoyuan Yi$^{5}$,\,   Wenhao Li$^{1,2,3}$,\,   Maosong Sun$^{1,2,3,4}$\thanks{\ \ Corresponding author. Email: sms@tsinghua.edu.cn}, \, Xing Xie$^{5}$ \\
    $^1$ Department of Computer Science and Technology, Tsinghua University, Beijing \\
    $^2$ Beijing National Research Center for Information Science and Technology \\
    $^3$ Institute for Artificial Intelligence, Tsinghua University, Beijing \\
    $^4$ Jiangsu Collaborative Innovation Center for Language Ability, Jiangsu Normal University, Xuzhou \\
    $^5$ Microsoft Research Asia \\
    \texttt{hu-jy21@mails.tsinghua.edu.cn,}\,\texttt{xiaoyuanyi@microsoft.com}
}

\begin{document}
\maketitle
\begin{abstract}
Variational Auto-Encoder (VAE) has been widely adopted in text generation. Among many variants, recurrent VAE learns token-wise latent variables with each conditioned on the preceding ones, which captures sequential variability better in the era of RNN. However, it is unclear how to incorporate such recurrent dynamics into the recently dominant Transformer due to its parallelism. In this work, we propose \modelname, a Transformer-based recurrent VAE structure. \modelname imposes recurrence on segment-wise latent variables with arbitrarily separated text segments and constructs the posterior distribution with residual parameterization. Besides, we design an acceleration method by approximating idempotent matrices, which allows parallelism while maintaining the conditional dependence of latent variables. We demonstrate that \modelname could enhance the entanglement of each segment and preceding latent variables and deduce a non-zero lower bound of the KL term, providing a theoretical guarantee of generation diversity. Experiments on various generation tasks show that \modelname achieves significantly improved diversity while maintaining satisfactory generation quality.
\end{abstract}

\section{Introduction}
\label{sec:1}
Variational Auto-Encoder (VAE)~\citep{kingma2013auto, rezende2014stochastic} has thrived in various text generation tasks due to its ability to learn flexible representations, such as machine translation~\cite{shah2018generative} and the generation of dialogue~\cite{zhao-etal-2017-learning}, story~\cite{MengHsuanYu2020DraftAE} and poetry~\cite{Mixpoet:20}. To further improve expressive power, diverse  VAE variants have been proposed. For example, IWAE~\cite{YuriBurda2016ImportanceWA} optimizes a tighter lower bound; Ladder VAE~\cite{CasperKaaeSnderby2016LadderVA} learns hierarchical latent representations and vMF-VAE~\cite{xu-durrett-2018-spherical} replaces Gaussian distributions with von Mises-Fisher distributions. 

Among all variants, temporal VAE \cite{fabius2015variational, chung2015recurrent} was prevalent in the era of RNN, which captures temporal variability by introducing the dependency of a series of latent variables with each associated with one time step. Such a VAE variant has succeeded in kinds of sequence modeling tasks, \textit{e.g.}, dialog generation~\cite{ByeongchangKim2020SequentialLK}, audio generation \cite{pmlr-v119-franceschi20a}, and time series prediction~\cite{LongyuanLi2019LearningID}.


Temporal VAE can be categorized into three paradigms according to the dependency of prior distributions at each time step: a) independent normal distributions (abbr. \emph{IND})~\cite{li-etal-2020-improving-variational}, b) context-conditioned Gaussian distributions (abbr. \emph{CGD})~\cite{du-etal-2018-variational} which are conditioned on preceding text, and c) recurrent Gaussian distributions (abbr. \emph{RGD}), \textit{i.e.}, Recurrent VAE~\cite{chung2015recurrent}, which are conditioned on preceding both text and latent variables\footnote{See Sec.~\ref{sec:3.1} for mathematical details of these paradigms.}. Both IND and CGD ignore the interaction of latent variables, limiting their expressive ability. In comparison, by introducing the dependency of latent variables, RGD could better model the sequential variability and thus greatly improve generation diversity while maintaining satisfactory quality. We provide the theoretical proof of such an advantage in Sec.~\ref{sec:4.3}.

These paradigms can be easily implemented with RNN benefiting from RNN's natural recurrent structure. Stepping into the age of Transformer~\cite{vaswani2017attention}, it is promising to adapt temporal VAE to this popular architecture. \emph{IND} and \emph{CGD} paradigms are naturally compatible with Transformer because their latent variables at each time step are independent which could be simply combined with the parallel computation of Transformer self-attention via causal and non-causal masks~\cite{lin2020variational}. However, there are no off-the-shelf solutions to incorporate \emph{RGD} into Transformer-based VAEs, since recurrence would be a natural obstacle to parallelism (recurrent latent variables need to be sequentially sampled), which limits the capacity of this potential VAE paradigm. 

\emph{Could we equip Transformer with such recurrent dynamics for better diversity while keeping the training parallelism?} To answer this question, we propose \modelname\footnote{\ \modelname: \textbf{T}ransformer \textbf{R}ecurrent \textbf{A}utoen\textbf{C}od\textbf{E}r}, a novel Transformer-based recurrent VAE structure. \modelname imposes recurrence on segment-wise (instead of token-wise) latent variables with arbitrary segmentation, \textit{e.g.}, sentences or segments with specified length. Besides, we construct the posterior distribution using residual parameterization and layer normalization, which could deduce a non-zero lower bound of the KL loss to alleviate KL vanishing~\cite{bowman-etal-2016-generating}. Moreover, to accelerate training, we design a method to recover the parallelism in Transformer by approximating idempotent parameter matrices for the latent space, leading to improved diversity, satisfactory quality, and faster training.

In summary, our contributions are as follows:
We are the first to ($i$) incorporate recurrent VAE into Transformer with recurrence on segment-wise latent variables which allows a flexible trade-off of diversity and quality; ($ii$) propose a method to recover parallelism and accelerate training with comparable performance; ($iii$) mathematically demonstrate that our model has a  deducted non-zero lower bound to mitigate KL vanishing, and a theoretical interpretation of diversity improvement. ($iv$) We validate the effectiveness of our model on two unconditional and one conditional generation tasks.

\section{Related Work}
VAE has shown great effectiveness in a wide range of NLG tasks, such as storytelling~\cite{MengHsuanYu2020DraftAE, LeFang2021TransformerbasedCV}, dialogue generation~\cite{serban2017hierarchical, bao-etal-2020-plato} and poetry composition~\cite{yi2021text}. To further improve the expressive ability of VAE, researchers propose various variants, \textit{e.g.}, vMF-VAE~\cite{xu-durrett-2018-spherical} that replaces the latent distribution with von Mises-Fisher distribution, \textit{ml}-VAE \cite{bouchacourt2018multi} that learns multi-level latent variables, and BN-VAE~\cite{zhu-etal-2020-batch} that utilizes batch normalization to get a non-zero KL lower bound.

Among all variants, temporal VAE is the most prevalent one in the era of RNN, which introduces latent variables at each timestep and could naturally fit with the recurrent structure of RNN. Existing temporal VAE fall into three paradigms according to the parameterization and dependence of the latent variables' prior distributions, namely \emph{IND}, \emph{CGD}, and \emph{RGD}, as mentioned in Sec. \ref{sec:1}. For example,  TWR-VAE~\cite{li-etal-2020-improving-variational} utilizes a timestep-wise regularisation through independent latent variables with IND. VAD~\cite{du-etal-2018-variational} incorporates CGD into latent variables and augments the posterior distribution with a backward RNN. Recurrent VAE~\cite{chung2015recurrent} learns token-wise latent variables with each sequentially conditioned on preceding ones as well as the context (\textit{i.e.}, RGD). By modeling the trajectory of both observed text sequences and latent space, recurrent VAE could capture the sequential variability better~\cite{NIPS2017_900c563b,hajiramezanali2020semi}. Besides, we will show that such recurrent dynamics could theoretically reinforce the dependence on the stochastic and generalized latent space, thus boosting generation diversity by a large margin.

Recently, with the flourishing of the powerful Transformer architecture, researchers have devoted to combining it with VAE for text modeling and generation~\cite{wang2019t, li-etal-2020-optimus, LeFang2021TransformerbasedCV, hu-etal-2022-fuse}. VAEs could promote generation diversity with satisfactory quality, benefiting from the intrinsic randomness in latent space. Therefore, VAE-based Transformers are essential for various tasks demanding creativity, such as advertising text generation \cite{shao-etal-2019-long}. Two of the temporal VAE paradigms, IND and CGD, can be easily adapted into Transformer. For instance, SVT~\cite{lin2020variational} applies CGD-based VAE to dialogue generation. Nonetheless, the integration of recurrent VAE is still an open challenge due to the conflict in the parallelism in Transformer and recurrent dependence of recurrent VAE. To fully exploit the expressive power of recurrence, we revisit recurrent VAE in Transformer and propose \modelname which possesses advantages of both generation diversity and training parallelism.
\section{Preliminaries}
\subsection{VAE}
As one of the representative generative models, VAE has proven to be an effective paradigm for estimating the data distribution by introducing a latent variable $\boldsymbol{z}$ and modeling the joint distribution:
\begin{equation}
    p(\boldsymbol{x}, \boldsymbol{z}) = p(\boldsymbol{x}|\boldsymbol{z})p(\boldsymbol{z}).
\end{equation}
The prior distribution $p(\boldsymbol{z})$ is commonly a standard Gaussian distribution. The conditional distribution $p(\boldsymbol{x}|\boldsymbol{z})$ is generally parameterized by a neural network, known as the \emph{generative network} (decoder) to recover the observed data from latent variables. Directly estimating $p(\boldsymbol{x}|\boldsymbol{z})$ brings the intractable posterior distribution $p(\boldsymbol{z}|\boldsymbol{x})$. Instead, VAE introduces a variational approximation $q(\boldsymbol{z}|\boldsymbol{x})$ and derives the Evidence Lower BOund (ELBO):
\begin{align}
    \begin{split}
    &\log p(\boldsymbol{x}) \ge \mathcal{L}_{ELBO}(\boldsymbol{x})=\\
    &\mathbb{E}_{q(\boldsymbol{z}|\boldsymbol{x})}[\log p(\boldsymbol{x}|\boldsymbol{z})] \!-\! \mathrm{KL}(q(\boldsymbol{z}|\boldsymbol{x}) || p(\boldsymbol{z})),
    \end{split}
\label{eq:elbo}
\end{align}
where $\mathrm{KL}$ means the Kullback-Leibler divergence.

In practice, the approximated posterior $q(\boldsymbol{z}|\boldsymbol{x})$ is parameterized as Gaussian distribution $\mathcal{N}(\boldsymbol{\mu}, \mathrm{diag}(\boldsymbol{\sigma^2}))$, where $\boldsymbol{\mu}$ and $\boldsymbol{\sigma}$ are estimated by a neural network, known as the \emph{inference network} (encoder). The generative network $p(\boldsymbol{x}|\boldsymbol{z})$ and inference network $q(\boldsymbol{z}|\boldsymbol{x})$ are jointly optimized by maximizing the lower bound in Eq.\eqref{eq:elbo}.

\subsection{Temporal VAE}
\label{sec:3.1}
Unlike standard VAE, which only involves one latent variable $\boldsymbol{z}$, temporal VAE learns one latent variable at each time step. Denote $\boldsymbol{z}_t \in \mathbb{R}^l$ and $\boldsymbol{x}_t \in \mathbb{R}^h$ as the latent variables and the observed data at $t$-th step, respectively. Next, we will present the mathematical details of three paradigms of temporal VAE, namely IND, CGD, and RGD. 

\paragraph{IND:} The prior distribution $p(\boldsymbol{z}_t)$ follows the standard Gaussian distribution $\mathcal{N}(\boldsymbol{0},\boldsymbol{I})$, and the posterior one conditions on the preceding context as $q(\boldsymbol{z}_t|\boldsymbol{x}_{\le t})$. Then, we  obtain the ELBO of IND:
\begin{equation}
\begin{aligned}
      &\sum\limits_{t=1}^{T}\mathbb{E}_{q(\boldsymbol{z}_t|\boldsymbol{x}_{\le t})}\log p(\boldsymbol{x}_{t}|\boldsymbol{z}_t,\! \boldsymbol{x}_{<t}))\\
      &-\mathrm{KL}(q(\boldsymbol{z}_t|\boldsymbol{x}_{\le t})||p(\boldsymbol{z}_t)).
\end{aligned}
\end{equation}

\paragraph{CGD:} CGD constructs the prior distribution considering the observed text $p(\boldsymbol{z}_t|\boldsymbol{x}_{<t})$ and the posterior one based on the complete text $\boldsymbol{x}=\{\boldsymbol{x}_1, \cdots, \boldsymbol{x}_{T}\}$. The lower bound of CGD is:
\begin{equation}
\begin{aligned}
\begin{aligned}
      &\sum\limits_{t=1}^{T}\mathbb{E}_{q(\boldsymbol{z}_t|\boldsymbol{x})}\log p(\boldsymbol{x}_{t}|\boldsymbol{z}_t, \boldsymbol{x}_{<t}))\\
       &-\mathrm{KL}(q(\boldsymbol{z}_t|\boldsymbol{x})||p(\boldsymbol{z}_t|\boldsymbol{x}_{<t})).
\end{aligned}
\end{aligned}
\end{equation}

\paragraph{RGD:} RGD parameterizes the generative process by the following factorization:
\begin{equation}
    p(\boldsymbol{x}_{\le T}, \boldsymbol{z}_{\le T}) = \prod\limits_{t=1}^{T}p(\boldsymbol{x}_{t}|\boldsymbol{z}_{\le t}, \boldsymbol{x}_{<t})p(\boldsymbol{z}_{t}|\boldsymbol{z}_{<t}, \boldsymbol{x}_{<t}).
\end{equation}
The latent variables $\boldsymbol{z}_t$ follows the prior distribution $p(\boldsymbol{z}_t|\boldsymbol{z}_{<t}, \boldsymbol{x}_{<t})$ and the posterior one follows $q(\boldsymbol{z}_t|\boldsymbol{z}_{<t}, \boldsymbol{x}_{\le t})$. Then, we obtain the ELBO:
\begin{equation}
\begin{aligned}
      &\mathbb{E}_{q(\boldsymbol{z}_{\le T}|\boldsymbol{x}_{\le T})}\Big[\sum\limits_{t=1}^{T}\log p(\boldsymbol{x}_{t}|\boldsymbol{z}_{\le t}, \boldsymbol{x}_{<t})\\
      &-\mathrm{KL}(q(\boldsymbol{z}_{t}|\boldsymbol{z}_{<t}, \boldsymbol{x}_{\le t})||p(\boldsymbol{z}_{t}|\boldsymbol{z}_{<t}, \boldsymbol{x}_{<t}))\Big],
\end{aligned}
\label{eq:loss}
\end{equation}
where $q(\boldsymbol{z}_{\le T}|\boldsymbol{x}_{\le T})$ can be factorized as:
\begin{equation}
    q(\boldsymbol{z}_{\le T}|\boldsymbol{x}_{\le T}) = \prod\limits_{t=1}^{T}q(\boldsymbol{z}_{t}|\boldsymbol{z}_{<t}, \boldsymbol{x}_{\le t}).
\end{equation}
We present the detailed deduction of Eq.\eqref{eq:loss} in Appendix \ref{proof_rgd}.

In an RNN-like backbone, we can construct the representation of $\boldsymbol{x}_{\le t}$ with the hidden states at $t$-th step and compute the distribution parameters of $\boldsymbol{z}_{t}$. 
\section{Method}
To incorporate recurrent VAE (RGD) into Transformer, we propose \modelname that learns recurrent segment-wise latent variables and design an acceleration method to make full use of the advantage of parallelism in Transformer. We present the adaption of recurrent VAE to Transformer and residual parameterization in Sec.~\ref{sec:4.1}, demonstrate the parallel training method in Sec.~\ref{sec:4.2}, and provide a theoretical interpretation of \modelname's effectiveness for boosting diversity in Sec. \ref{sec:4.3}.
\begin{figure*}[t]
\centering
\includegraphics[scale=0.51]{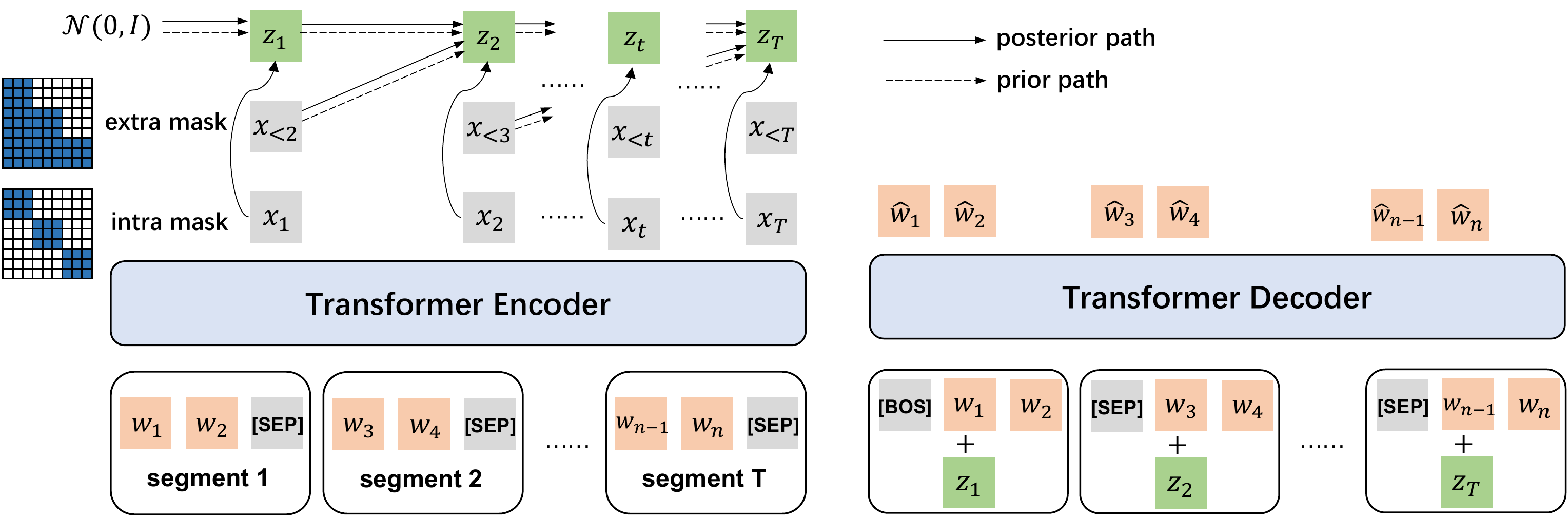} \
\caption{Architecture of \modelname. We add a special token at the end of each segment and obtain the representation of $\boldsymbol{x}_t$ and $\boldsymbol{x}_{<t}$ from the Transformer encoder (inference network) with two kinds of modified attention mask matrix. The solid and dotted lines are the posterior and prior paths, respectively. The sampled latent variables are added to the token embedding in the Transformer decoder (generator network).}
\label{fig:model}
\end{figure*}
\subsection{Transformer-based Recurrent VAE}
\label{sec:4.1}
Different from the token-wise latent variables implemented in RNN-based VAEs, \modelname learns segment-wise $\boldsymbol{z}_t$ based on the representation of $t$-th segment, $\boldsymbol{x}_t$. We can devise different principles to separate the segments, such as the inherent separation like sentence or utterance, or specifying a fixed segment length. We add a special token \texttt{[SEP]} to the end of each segment.

Fig.~\ref{fig:model} depicts the architecture of \modelname. At the encoder, we design two kinds of attention mask matrices. First, we introduce an \textit{extra mask matrix}, a partitioned lower triangular matrix (the left of Fig.~\ref{fig:model}), which allows each token to attend to all tokens in the same segment and previous segments. Second, we design an \textit{intra mask matrix}, a strict partitioned matrix to make each token attend to only the tokens within the same segment. We input the separated text sequence into the Transformer encoder twice, with the extra and intra mask matrix, respectively. Then, the output of $t$-th \texttt{[SEP]} from the final encoder layer can be used as the representation of $\boldsymbol{x}_{<t}$ and $\boldsymbol{x}_{t}$. Now, we can obtain the parameters of the prior distribution of $\boldsymbol{z}_t$ by:
\begin{align}
    p(\boldsymbol{z}_t|\boldsymbol{z}_{<t}, \boldsymbol{x}_{<t}) &= \mathcal{N}(\boldsymbol{\mu}_t, \mathrm{diag}(\boldsymbol{\sigma}^2_t)),  \\
    [\boldsymbol{\mu}_t, \log\boldsymbol{\sigma}^2_t] &= f(\boldsymbol{z}_{t-1}, \boldsymbol{x}_{<t}),
\end{align}
where $f$ is the prior network, parameterized as linear layers $\boldsymbol{W}^f_{\boldsymbol{\mu}}, \boldsymbol{W}^f_{\boldsymbol{\sigma}} \in \mathbb{R}^{(l+h)\times l}$. The prior distribution of $\boldsymbol{z}_1$ is the standard Gaussian distribution.

For the posterior distribution, we utilize residual parameterization \cite{ArashVahdat2020NVAEAD} that parameterizes the relative difference between prior and posterior distributions. In this case, the difference lies in $\boldsymbol{x}_t$. Therefore, we can compute the posterior distribution as:
\begin{align}
\label{eq:9}
    q(\boldsymbol{z}_t|\boldsymbol{z}_{<t}, \boldsymbol{x}_{\le t}) &= \mathcal{N}(\boldsymbol{\mu}_t\!+\!\Delta\boldsymbol{\mu}_t, \mathrm{diag}(\boldsymbol{\sigma}^2_t\!\cdot\!\Delta\boldsymbol{\sigma}^2_t)),  \\
    [\Delta\boldsymbol{\mu}_t, \log\Delta\boldsymbol{\sigma}^2_t] &= \mathrm{LayerNorm}(g(\boldsymbol{x}_t)),
\end{align}
where $g$ is the posterior network, parameterized as $\boldsymbol{W}^g_{\boldsymbol{\mu}}, \boldsymbol{W}^g_{\boldsymbol{\sigma}} \in \mathbb{R}^{h\times l}$. We regulate the output of $g$ by layer normalization \cite{JimmyBa2016LayerN}. To reveal the benefits of residual parameterization and layer normalization, we give the following theorem:
\begin{theorem}
The expectation of the KL term $\mathrm{KL}(q(\boldsymbol{z}_{t}|\boldsymbol{z}_{<t}, \boldsymbol{x}_{\le t})||p(\boldsymbol{z}_{t}|\boldsymbol{z}_{<t}, \boldsymbol{x}_{<t}))$ has a lower bound $\frac{l(\gamma^2 + \beta^2)}{2\boldsymbol{\sigma}_t^2}$, where $l$ is the latent dimension, $\gamma$ and $\beta$ are the parameters of layer normalization. 
\label{thm1}
\end{theorem}

We leave the proof in Appendix \ref{proof2}. Theorem \ref{thm1} indicates that we can easily control the lower bound of the KL term by setting a fixed $\gamma$ and hence mitigate KL vanishing. We choose layer normalization here since it is superior to batch normalization in Transformer based models~\cite{shen2020powernorm} (See Table~\ref{tab:ablation}). Besides, Theorem \ref{thm1} is compatible with both unconditional and conditional generation compared to the BN VAE model~\cite{zhu-etal-2020-batch}.

After deriving the prior and posterior distributions, we can compute the KL loss and sample the latent variables with the reparameterization trick \cite{kingma2013auto}. The sampled latent variables $\boldsymbol{z}_t$ are injected into the Transformer decoder by adding with the input embedding. 

\subsection{Parallel Training for Recurrent VAE}
\label{sec:4.2}
The method above requires sequentially sampling each latent variable, which hinders the parallelism training in Transformer. For acceleration, we further design a parallel training method. 

With the reparameterization trick, when sampling $\boldsymbol{z}_t \sim \mathcal{N}(\boldsymbol{\mu}, \boldsymbol{\sigma}^2)$, we actually first sample a white noise $\boldsymbol{\epsilon} \sim N(\boldsymbol{0}, \boldsymbol{I})$, then get $\boldsymbol{z} \!=\! \boldsymbol{\mu} + \boldsymbol{\epsilon}\circ\boldsymbol{\sigma}$, where $\circ$ is element-wise multiplication. We omit LayerNorm for simplicity. For each $t$, we have:
\begin{align}
\label{eq:13}
    \boldsymbol{z}_t\! \approx \boldsymbol{v}_t + \sum\limits_{i=1}^{t-1}\boldsymbol{u}_i,
\end{align}
\begin{equation}
    \begin{aligned}
         \boldsymbol{v}_t=&(\boldsymbol{W}^f_{\boldsymbol{\mu}2}\boldsymbol{x}_{<t}+ \boldsymbol{W}^g_{\boldsymbol{\mu}}\boldsymbol{x}_{t})+\boldsymbol{\epsilon}_t+ \\
         &\Big(\boldsymbol{W}^f_{\boldsymbol{\sigma}2}\boldsymbol{x}_{<t}+\boldsymbol{W}^g_{\boldsymbol{\sigma}}\boldsymbol{x}_{t}\Big)\!\circ\!\boldsymbol{\epsilon}_t, 
    \end{aligned}
\end{equation}
\begin{equation}
    \boldsymbol{u}_t = \boldsymbol{W}^f_{\boldsymbol{\mu}1}\boldsymbol{v}_{t} + \boldsymbol{W}^f_{\boldsymbol{\sigma}1}\boldsymbol{v}_{t}\circ\boldsymbol{\xi}_t,
\end{equation}
where $\boldsymbol{\epsilon}_t$ and $\boldsymbol{\xi}_i$ are independent white noises sampled from standard Gaussian distribution. $[\boldsymbol{W}^f_{\boldsymbol{\mu}1}, \boldsymbol{W}^f_{\boldsymbol{\mu}2}], [\boldsymbol{W}^f_{\boldsymbol{\sigma}1}, \boldsymbol{W}^f_{\boldsymbol{\sigma}2}]$ are the split of parameter matrices of function $f$ with $\boldsymbol{W}^f_{\boldsymbol{\mu}1}, \boldsymbol{W}^f_{\boldsymbol{\sigma}1} \in \mathbb{R}^{l\times l}$ and $\boldsymbol{W}^f_{\boldsymbol{\mu}2}, \boldsymbol{W}^f_{\boldsymbol{\sigma}2} \in \mathbb{R}^{h\times l}$. We leave the complete derivation in Appendix \ref{proof1}.

In this way, we can parallelly train the model while keeping the advantage of RGD. We can parallelly compute $\boldsymbol{v}_t$ for all time steps first, and then obtain $\boldsymbol{u}_t$ in parallel based on $\boldsymbol{v}_t$. Then we can make the matrix multiplication between the concatenation of $\boldsymbol{u}_t$ of all time steps and a lower triangular matrix of ones to parallelly obtain $\sum_{i=1}^{t-1}\boldsymbol{u}_i$. Finally, by making a sum, we get the approximation of $\boldsymbol{z}_t$. Similarly, we obtain the latent samples from the posterior distribution in Eq.\eqref{eq:9}.

In the approximation of Eq.\eqref{eq:13}, we assume that $\boldsymbol{W}^f_{\boldsymbol{\mu}1}$ and $\boldsymbol{W}^f_{\boldsymbol{\sigma}1}$ are idempotent matrices to avoid the power of matrix. However, such simplification may bring too much noise. To stabilize the training process, we adopt spectral normalization~\cite{miyato2018spectral} to restrain $\boldsymbol{W}^f_{\boldsymbol{\mu}1}$ and $\boldsymbol{W}^f_{\boldsymbol{\sigma}1}$.
\subsection{Why Could \modelname Boost Diversity}
\label{sec:4.3}
We give a theoretical demonstration of the advantage of \modelname on the improvement of generation diversity. We have the following theorem:
\begin{theorem}
Each reconstruction term in the ELBO, $\mathbb{E}_{q(\boldsymbol{z}_{\le t}|\boldsymbol{x}_{\le t})}p(\boldsymbol{x}_t|\boldsymbol{z}_{\le t}, \boldsymbol{x}_{< t})$, is upper bounded by $I_q(\boldsymbol{x}_t;\boldsymbol{z}_{\le t}|\boldsymbol{x}_{<t})$, where $I$ is the mutual information. 
\label{thm2}
\end{theorem}
The proof can be found in Appendix \ref{proof3}.
Based on Theorem \ref{thm2}, optimizing the reconstruction terms means maximizing $I(\boldsymbol{x}_t;\boldsymbol{z}_{\le t}|\boldsymbol{x}_{<t})$, which could enhance the dependency between $\boldsymbol{x}_t$ and $\boldsymbol{z}_{\le t}$. In this way, the model would rely more on the flexible latent space than the deterministic context text, bringing more randomness to improve diversity while maintaining satisfactory coherence.
\section{Experiment}

\subsection{Dataset}
We carry out experiments on two datasets for language modeling and unconditional generation, including the Yelp and Yahoo \citep{yang2017improved, JunxianHe2019LaggingIN} and one dataset, WritingPrompts (WP) \citep{fan-etal-2018-hierarchical} for conditional generation. We list the detailed data statistics of these datasets in Table \ref{tab:stat_dataset}. Due to the limited computation capability, we restrain a max length of 750 for training in WP. 
\subsection{Implementation Details}
We use the pretrained language model GPT-2 \cite{Radford2019LanguageMA} as the backbone. The encoder and decoder of \modelname share the same parameters initialized with GPT-2 and are fine-tuned on the target datasets. We choose 32 for the dimension of latent space and use the cyclical annealing trick~\cite{fu-etal-2019-cyclical} during training. We set batch size as 32 and the learning rate as $5e-5$, $\gamma$ in the layer normalization as 3. We separate segments with fixed length 10 on Yelp and Yahoo datasets and use the initial sentence as a segment on the WP dataset with NLTK toolkit \cite{bird-2006-nltk} for segmenting. We use the top-k sampling strategy \cite{AriHoltzman2020TheCC} to decode the sequence with k as $50$ for all models on all datasets. When the generated token is \texttt{[SEP]}, we sample a new latent variable based on the prior distribution to enter a new segment. We implement \modelname and other VAE baselines with open-source Huggingface Transformers \citep{wolf-etal-2020-transformers} library of v4.10.0 and use NVIDIA GeForce RTX 3090 to conduct all experiments. 
\begin{table}[t]
\centering
\scalebox{0.87}{
\begin{tabular}{l|c|c|c|c}

\toprule
Dataset        & \# Train     & \# Dev   & \# Test   & Length      \\ \toprule
Yelp           & 100k         & 10k      & 10k       &     96        \\ 
Yahoo          & 100k         & 10k      & 10k       &     79        \\  \midrule
WP             & 164k         & 15k      & 15k       &  421       \\  \bottomrule
\end{tabular}
}
\caption{\textbf{Statistics of datasets}. Length means the average text length of the three datasets.}
\label{tab:stat_dataset}
\end{table}

\subsection{Baseline}
We compare \modelname with several transformer-based solid models. All baseline models are utilized with the same backbone model.

\textbf{GPT-2}: We finetune the GPT-2 model~\cite{Radford2019LanguageMA} on each dataset.

\textbf{IND/CGD}: We implement the Transformer version of the two models, TWR-VAE~\cite{li-etal-2020-improving-variational} and VAD~\cite{du-etal-2018-variational} which belongs to IND and CGD respectively. The segment separation is consistent with \modelname.

\textbf{No Recurrence} \cite{li-etal-2020-optimus}: We remove the recurrence and only involve one latent variable to verify the effectiveness of recurrence. The latent variables are injected into the decoder by being added with the text embedding in the decoder. 

\subsection{Metrics}
We evaluate unconditional generation tasks with three perspectives; \textbf{(a) Representation Learning}: we report ELBO, KL, mutual information (MI)~\citep{AlexanderAAlemi2016DeepVI} and activate units (AU)~\citep{YuriBurda2016ImportanceWA}. We change the threshold to 0.1 when computing the AU metric to distinguish different models further. \textbf{(b) Generation Quality}: we report PPL and CND~\citep{JianingLi2020OnTR} to evaluate the generation capacity of models. Different from standard auto-regressive language models like GPT-2, VAE-based models could not estimate exact PPL. Therefore, following~\citet{JunxianHe2019LaggingIN}, we use importance-weighted samples to approximate $\log p(\boldsymbol{x})$ and estimate PPL. CND measures the divergence between the generated one and the ground truth testset. \textbf{(c) Generation Diversity}: we report Self-BLEU \citep{YaomingZhu2018TexygenAB}, Dist \citep{li-etal-2016-diversity} and JS (Jaccard similarity) \citep{KeWang2018SentiGANGS} to evaluate the diversity of generated text. 

We consider the quality and diversity in story generation. We report BLEU~\cite{papineni-etal-2002-bleu}, Rouge-1, 2, L \citep{lin-hovy-2002-manual}, and BERTScore \citep{TianyiZhang2020BERTScoreET} to evaluate the quality of generated samples, and the same diversity metrics used in unconditional generation. More details about metrics are listed in Appendix \ref{apx_sec:metrics}.

\subsection{Results}
\begin{table*}[htp]
\centering
\scalebox{0.9}{
\begin{tabular}{cccccccccc}
\toprule
\multicolumn{1}{c|}{\multirow{2}{*}{Model}} & \multicolumn{4}{c|}{Representation   Learning} & \multicolumn{2}{c|}{Generation Quality} & \multicolumn{3}{c}{Generation   Diversity} \\ \cline{2-10} 
\multicolumn{1}{c|}{}  &\multicolumn{1}{c|}{ELBO$\downarrow$}   & \multicolumn{1}{c|}{KL$\uparrow$}    & \multicolumn{1}{c|}{MI$\uparrow$}    & \multicolumn{1}{c|}{AU$\uparrow$} & \multicolumn{1}{c|}{PPL$\downarrow$}  & \multicolumn{1}{c|}{CND$\downarrow$} &  \multicolumn{1}{c|}{SB$\downarrow$}      & \multicolumn{1}{c|}{Dist$\uparrow$} & JS $\downarrow$     \\ \midrule
\multicolumn{10}{c}{Dataset: Yelp} \\ \midrule
\multicolumn{1}{c|}{GPT-2}                   & \multicolumn{1}{c|}{-}  & \multicolumn{1}{c|}{-}       & \multicolumn{1}{c|}{-}      & \multicolumn{1}{c|}{-}      & \multicolumn{1}{c|}{22.13}   &  \multicolumn{1}{c|}{0.68}   & \multicolumn{1}{c|}{65.90}  & \multicolumn{1}{c|}{15.96}      &       0.51       \\ 
\multicolumn{1}{c|}{IND}   &  \multicolumn{1}{c|}{328.17} & \multicolumn{1}{c|}{2.32}  & \multicolumn{1}{c|}{2.92}  & \multicolumn{1}{c|}{25}    & \multicolumn{1}{c|}{15.35}   & \multicolumn{1}{c|}{0.76}  & \multicolumn{1}{c|}{61.24}   & \multicolumn{1}{c|}{16.24}     &  0.45            \\
\multicolumn{1}{c|}{CGD}  & \multicolumn{1}{c|}{327.84}  & \multicolumn{1}{c|}{0.07}   & \multicolumn{1}{c|}{0.04}   & \multicolumn{1}{c|}{2}   & \multicolumn{1}{c|}{16.29}& \multicolumn{1}{c|}{\underline{0.46}}  & \multicolumn{1}{c|}{68.36}    & \multicolumn{1}{c|}{13.68}    &    0.59        \\ 
\multicolumn{1}{c|}{No Recurrence}  & \multicolumn{1}{c|}{327.35}  & \multicolumn{1}{c|}{\underline{3.89}}   & \multicolumn{1}{c|}{3.88}   & \multicolumn{1}{c|}{\underline{26}}   & \multicolumn{1}{c|}{20.12}& \multicolumn{1}{c|}{\textbf{0.43}}  & \multicolumn{1}{c|}{65.38}    & \multicolumn{1}{c|}{15.47}    &    0.44          \\ \midrule
\multicolumn{1}{c|}{\modelname-R}                  & \multicolumn{1}{c|}{\textbf{298.61}} & \multicolumn{1}{c|}{3.51} & \multicolumn{1}{c|}{\textbf{19.57}} & \multicolumn{1}{c|}{\textbf{32}} & \multicolumn{1}{c|}{\textbf{13.03}}   & \multicolumn{1}{c|}{0.47} & \multicolumn{1}{c|}{\textbf{57.02}}  & \multicolumn{1}{c|}{\textbf{23.01}}  & \textbf{0.33}   \\ 
\multicolumn{1}{c|}{\modelname-P}                  & \multicolumn{1}{c|}{\underline{315.28}} & \multicolumn{1}{c|}{\textbf{4.52}} & \multicolumn{1}{c|}{\underline{5.25}} & \multicolumn{1}{c|}{\textbf{32}} & \multicolumn{1}{c|}{\underline{14.88}}   & \multicolumn{1}{c|}{0.82} & \multicolumn{1}{c|}{\underline{60.25}}  & \multicolumn{1}{c|}{\underline{16.58}}  & \underline{0.40} \\ \midrule

\multicolumn{10}{c}{Dataset: Yahoo}  \\ \midrule
\multicolumn{1}{c|}{GPT-2}                   & \multicolumn{1}{c|}{-}  & \multicolumn{1}{c|}{-}       & \multicolumn{1}{c|}{-}      & \multicolumn{1}{c|}{-}      & \multicolumn{1}{c|}{24.17}   &  \multicolumn{1}{c|}{0.55}   & \multicolumn{1}{c|}{54.06}  & \multicolumn{1}{c|}{\underline{21.07}}   &       \underline{0.28}       \\ 
\multicolumn{1}{c|}{IND}   &  \multicolumn{1}{c|}{285.61} & \multicolumn{1}{c|}{2.31}  & \multicolumn{1}{c|}{2.89}  & \multicolumn{1}{c|}{20}    & \multicolumn{1}{c|}{17.05}   & \multicolumn{1}{c|}{0.90}  & \multicolumn{1}{c|}{53.08}   & \multicolumn{1}{c|}{20.55}     &  0.45            \\
\multicolumn{1}{c|}{CGD}  & \multicolumn{1}{c|}{285.92}  & \multicolumn{1}{c|}{0.17}   & \multicolumn{1}{c|}{0.12}   & \multicolumn{1}{c|}{3}   & \multicolumn{1}{c|}{18.84}& \multicolumn{1}{c|}{\textbf{0.45}}  & \multicolumn{1}{c|}{58.81}    & \multicolumn{1}{c|}{18.12}    &    0.53        \\ 
\multicolumn{1}{c|}{No Recurrence}  & \multicolumn{1}{c|}{286.89}  & \multicolumn{1}{c|}{3.62}   & \multicolumn{1}{c|}{5.65}   & \multicolumn{1}{c|}{\underline{25}}   & \multicolumn{1}{c|}{21.18}& \multicolumn{1}{c|}{\textbf{0.45}}  & \multicolumn{1}{c|}{54.15}    & \multicolumn{1}{c|}{20.78}    &    0.32      \\ \midrule
\multicolumn{1}{c|}{\modelname-R}                  & \multicolumn{1}{c|}{\textbf{258.25}} & \multicolumn{1}{c|}{\underline{3.65}} & \multicolumn{1}{c|}{\textbf{43.20}} & \multicolumn{1}{c|}{\textbf{32}} & \multicolumn{1}{c|}{\textbf{14.56}}   & \multicolumn{1}{c|}{\underline{0.53}} & \multicolumn{1}{c|}{\textbf{47.33}}  & \multicolumn{1}{c|}{\textbf{26.35}}  & \textbf{0.26}   \\ 
\multicolumn{1}{c|}{\modelname-P}   & \multicolumn{1}{c|}{\underline{278.35}} & \multicolumn{1}{c|}{\textbf{3.99}} & \multicolumn{1}{c|}{\underline{8.72}} & \multicolumn{1}{c|}{\textbf{32}} & \multicolumn{1}{c|}{\underline{16.85}}   & \multicolumn{1}{c|}{0.78} & \multicolumn{1}{c|}{\underline{52.75}}  & \multicolumn{1}{c|}{20.61}  & 0.30   \\ \bottomrule
\end{tabular}
}
\caption{Evaluation results for unconditional generation. SB: Self-BLEU. \modelname-R: \modelname with standard RGD. \modelname-P: The parallel version of \modelname. The best/second best results are in \textbf{bold} and \underline{underlined}, respectively.}
\label{tab:uncond_task}
\end{table*}
\subsubsection{Unconditional Generation}
We present the results of the unconditional generation task on Yelp and Yahoo datasets in Table \ref{tab:uncond_task}. As the results show, \modelname achieves significant improvement on most metrics. Better ELBO and MI indicate \modelname has stronger capability of representation learning. Especially, considerably higher MI empirically validates Theorem \ref{thm2}, which manifests that with the RGD mechanism, the observed data will be connected to the latent space more closely. Higher KL and AU also empirically show the benefit of residual parameterization. Besides, lower PPL and comparable CND indicates acceptable quality of the text generated by \modelname.

Moreover, \modelname can produce much more diverse text compared with baselines. Among the baselines, the CGD baseline suffers from the KL vanishing problem, which means the decoder ignores the latent variables. Therefore, without the randomness arising from latent variables, CGD performs the worst on generation diversity. In contrast, \modelname obtains the most prominent enhancement on all diversity metrics. Such improvement originates from two aspects: First, compared to GPT-2 and No Recurrence, sampling latent variables at each time step brings extra randomness for the output. Second, Unlike the other two temporal VAE baselines, \modelname is endowed with the theoretical advantage of RGD, which strengthens the interaction between the text and latent space and then absorbs much flexibility from the generalized latent space. IND simply increases randomness from standard Gaussian prior distribution, which negatively affects quality while causing limited diversity.

Lastly, comparing \modelname-R and \modelname-P, despite the slight drop of quality and diversity, \modelname-P still outperforms baselines on most metrics except CND. Such marginal cost is acceptable considering the acceleration with parallelism. See the comparison of training speeds in Sec.~\ref{sec:analysis}. The performance loss is mainly caused by the last two approximation steps in Appendix~\ref{proof1}. Theoretically, we approximate the multiplication of $t$ independent random variables as one Gaussian distribution using the central limit theorem (holds for infinite length) in Eq.\eqref{eq:25}, and restrict two parameter matrices to be idempotent (holds when the eigenvalue is $1$ or $0$) in Eq.\eqref{eq:26}. In practice, these assumptions don’t hold because the sequence length is not infinite, and the spectral normalization only restrains the largest eigenvalue (strictly limiting makes optimization difficult). Therefore, empirically we hurt the interaction of latent variables and sacrifice some information in them, leading to decreased performance compared to \modelname-R.

\begin{table*}[htp]
\centering
\scalebox{0.95}{
\begin{tabular}{ccccccccc}
\toprule
\multicolumn{1}{c|}{\multirow{2}{*}{Model}} & \multicolumn{5}{c|}{Quality} & \multicolumn{3}{c}{Diversity}   \\ \cline{2-9} 
\multicolumn{1}{c|}{}                       & \multicolumn{1}{c|}{BLEU$\uparrow$} & \multicolumn{1}{c|}{Rouge-1$\uparrow$} & \multicolumn{1}{c|}{Rouge-2$\uparrow$} & \multicolumn{1}{c|}{Rouge-L$\uparrow$} & \multicolumn{1}{c|}{BertScore$\uparrow$} & \multicolumn{1}{c|}{SB$\downarrow$} & \multicolumn{1}{c|}{Dist$\uparrow$} & \multicolumn{1}{c}{JS $\downarrow$} \\ \midrule
\multicolumn{1}{c|}{GPT-2}       & \multicolumn{1}{c|}{27.89}     & \multicolumn{1}{c|}{27.72}        & \multicolumn{1}{c|}{3.92}        & \multicolumn{1}{c|}{10.18}        & \multicolumn{1}{c|}{78.12}          & \multicolumn{1}{c|}{53.78}   &    \multicolumn{1}{c|}{22.99}   &  0.51 \\ 
\multicolumn{1}{c|}{IND}        & \multicolumn{1}{c|}{31.17}     & \multicolumn{1}{c|}{32.44}      & \multicolumn{1}{c|}{4.35}        & \multicolumn{1}{c|}{11.39}      & \multicolumn{1}{c|}{\textbf{81.31}}          & \multicolumn{1}{c|}{67.44}      &    \multicolumn{1}{c|}{13.69}   &   1.34  \\ 
\multicolumn{1}{c|}{CGD}         & \multicolumn{1}{c|}{\textbf{31.65}}     & \multicolumn{1}{c|}{\textbf{32.56}}      & \multicolumn{1}{c|}{\textbf{4.46}}        & \multicolumn{1}{c|}{\textbf{11.59}}        & \multicolumn{1}{c|}{81.28}          & \multicolumn{1}{c|}{68.25}   &    \multicolumn{1}{c|}{14.87}   &   1.97  \\ 
\multicolumn{1}{c|}{No Recurrence}  & \multicolumn{1}{c|}{31.57}     & \multicolumn{1}{c|}{32.27}      & \multicolumn{1}{c|}{4.28}        & \multicolumn{1}{c|}{11.30}        & \multicolumn{1}{c|}{81.25}     & \multicolumn{1}{c|}{67.25}   &    \multicolumn{1}{c|}{13.51}   &    1.38  \\ \midrule
\multicolumn{1}{c|}{\modelname-R}  & \multicolumn{1}{c|}{30.65}     & \multicolumn{1}{c|}{31.80}        & \multicolumn{1}{c|}{4.12}        & \multicolumn{1}{c|}{11.15}        & \multicolumn{1}{c|}{81.14}          & \multicolumn{1}{c|}{64.11}   &    \multicolumn{1}{c|}{13.18}   & 0.84 \\ 
\multicolumn{1}{c|}{\modelname-P}  & \multicolumn{1}{c|}{28.55}     & \multicolumn{1}{c|}{29.52}  & \multicolumn{1}{c|}{3.48}        & \multicolumn{1}{c|}{10.98}       & \multicolumn{1}{c|}{80.63}  & \multicolumn{1}{c|}{\textbf{47.26}}   &    \multicolumn{1}{c|}{\textbf{26.17}}  &  \textbf{0.70} \\ \bottomrule
\end{tabular}}
\caption{Evaluation results for conditional generation.}
\label{tab:cond_task}
\end{table*}

\subsubsection{Conditional Generation}
We report the results of conditional generation on the WP dataset. As shown in Table~\ref{tab:cond_task}, \modelname achieves comparable generation performance on quality (still better than GPT-2) and significant improvement on diversity (especially on Self-Bleu and Dist compared with other VAE baselines). Although the quality of text generated by \modelname-P is relatively defective, \modelname-P still outperforms GPT-2 on both quality and diversity. The overall enhancement of diversity empirically further validates our theoretical analysis.

Interestingly, \modelname-P achieves better generation diversity than \modelname-R on WP, which is opposite to the results on both Yelp and Yahoo datasets. This contrary tendency of diversity mainly originates from different sequence lengths. On Yelp and Yahoo datasets, whose text are relatively short, \modelname-R performs better. However, on the WP dataset with much longer text, the approximation in \modelname-P reduces the exploitation of $x_{<t}$ by reducing the interaction of each $x_t$, as shown in Eq.\eqref{eq:26}, which loosens the dependency on context and forces \modelname-P to produce more uncertain text (better diversity but lower quality) than \modelname-R.

\begin{table}[t]
\centering
\scalebox{0.9}{
\begin{tabular}{c|c|c|c}
\toprule
Model             & Fluency     & Coherence & Novelty \\ \midrule
GPT2              &     2.18    &   2.28    &    \underline{2.46}      \\
\hline
IND               &     2.38    &   2.41    &   2.40      \\
CGD               &     \textbf{2.43}    &    \underline{2.47}    &   2.41      \\
No Recurrence     &     \underline{2.42}    &   \textbf{2.51}    &   2.44      \\ \midrule
\modelname-R      &     2.40    &   2.46    &   \textbf{2.55}   \\
\modelname-P      &     2.24    &   2.32    &    \underline{2.46}      \\ \bottomrule
\end{tabular}}
\caption{Human evaluation results on the WP dataset. The scores range from 1 (worst) to 5 (best). The p-value $<$ 0.01, and the Kappa score is 0.61, which means the evaluation is in an acceptable inter-annotator agreement.}
\label{tab:human_eval}
\end{table}

\subsection{Human Evaluation}
We conduct the human evaluation on the WP dataset. We generate 50 samples given the source input from the testset with each model and invite five proficient annotators to access the generated text by scoring three criteria: \textbf{Fluency} (whether generated text are syntactically fluent), \textbf{Coherence} (whether the generated part is consistently structured and coherent with input), and \textbf{Novelty} (whether each generated instance is novel and distinct), which cover both generation quality and quality we care about. See Appendix \ref{sec:hm_detail} for more evaluation details.

We report the evaluation results in Table \ref{tab:human_eval}. \modelname obtains satisfactory results on Fluency and Coherence but stands superior to other baselines on Novelty. The results of the human evaluation are consistent with automatic evaluations.
\begin{table}[htp]
\centering
\scalebox{0.84}{
\begin{tabular}{c|c|c|c|c|c}
\toprule
Model             &    ELBO $\downarrow$   &   CND$\downarrow$   &    SB$\downarrow$     & Dist$\uparrow$ &   JS$\downarrow$ \\ \midrule
\modelname-R      &     298.61  &   0.47  &   57.02   &  23.01   &   0.33     \\
-LN               &     307.33  &   0.49  &   64.89   &  15.37   &   0.43        \\ 
-RP               &     326.88  &   0.78  &   67.89   &  13.87   & 0.58         \\ 
+BN               &     305.02  &   0.63  &   59.25   &  22.88   &   0.35  \\ \midrule
\modelname-P      &     315.28  &   0.82  &   60.25   &  16.58   & 0.40  \\ 
-SN     &     318.25  &    0.91  &   62.32   &  17.22 & 0.43 \\ \bottomrule
\end{tabular}}
\caption{Ablation study on Yelp. +BN means replacing layer normalization with batch normalization. -LN, -RP and -SN means removing layer normalization, residual design and spectral normalization, respectively.}
\label{tab:ablation}
\end{table}
\subsection{Ablation Study}
Table \ref{tab:ablation} shows the results of the ablation study on the Yelp dataset. We mainly justify the effectiveness of residual parameterization, layer normalization used after the posterior network, and spectral normalization for the prior network in \modelname-P. We also compare batch normalization by replacing layer normalization with batch normalization. Experimental results show that all of them benefit the quality or diversity. Specifically, without layer normalization or residual parameterization, \modelname will tend to ignore the latent variables and loss the diversity brought by the recurrent structure during generation. The difference between batch normalization and layer normalization is relatively marginal, while the former still performs better.

\begin{figure*}[ht]
\centering
\begin{minipage}[t]{0.48\textwidth}
\centering
\includegraphics[scale=0.38]{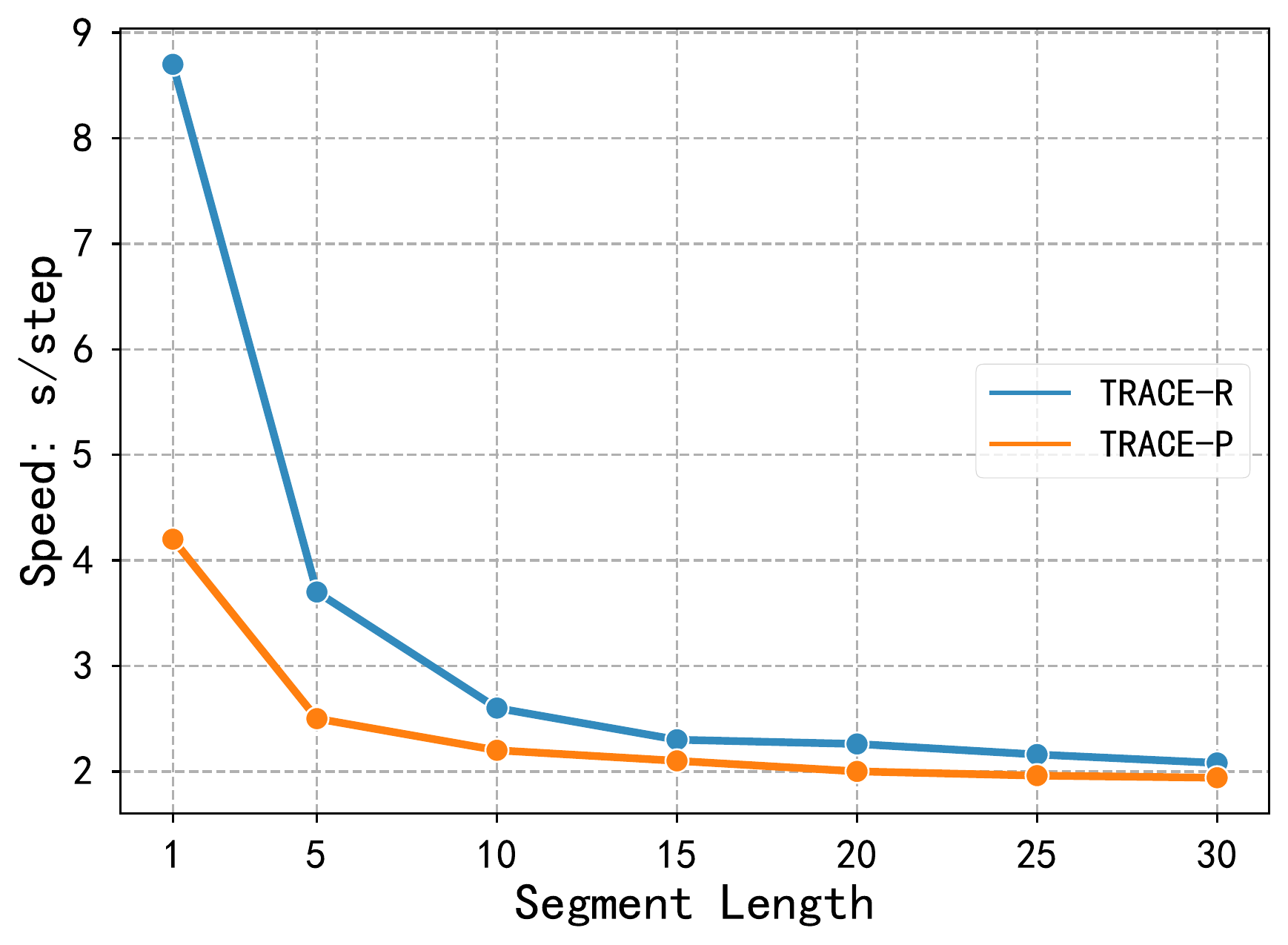}
\caption{Training speed (average seconds per step) of \modelname-R and \modelname-P with different segment lengths.}
\label{fig:speed}
\end{minipage}
\hfill
\begin{minipage}[t]{0.48\textwidth}
\includegraphics[scale=0.38]{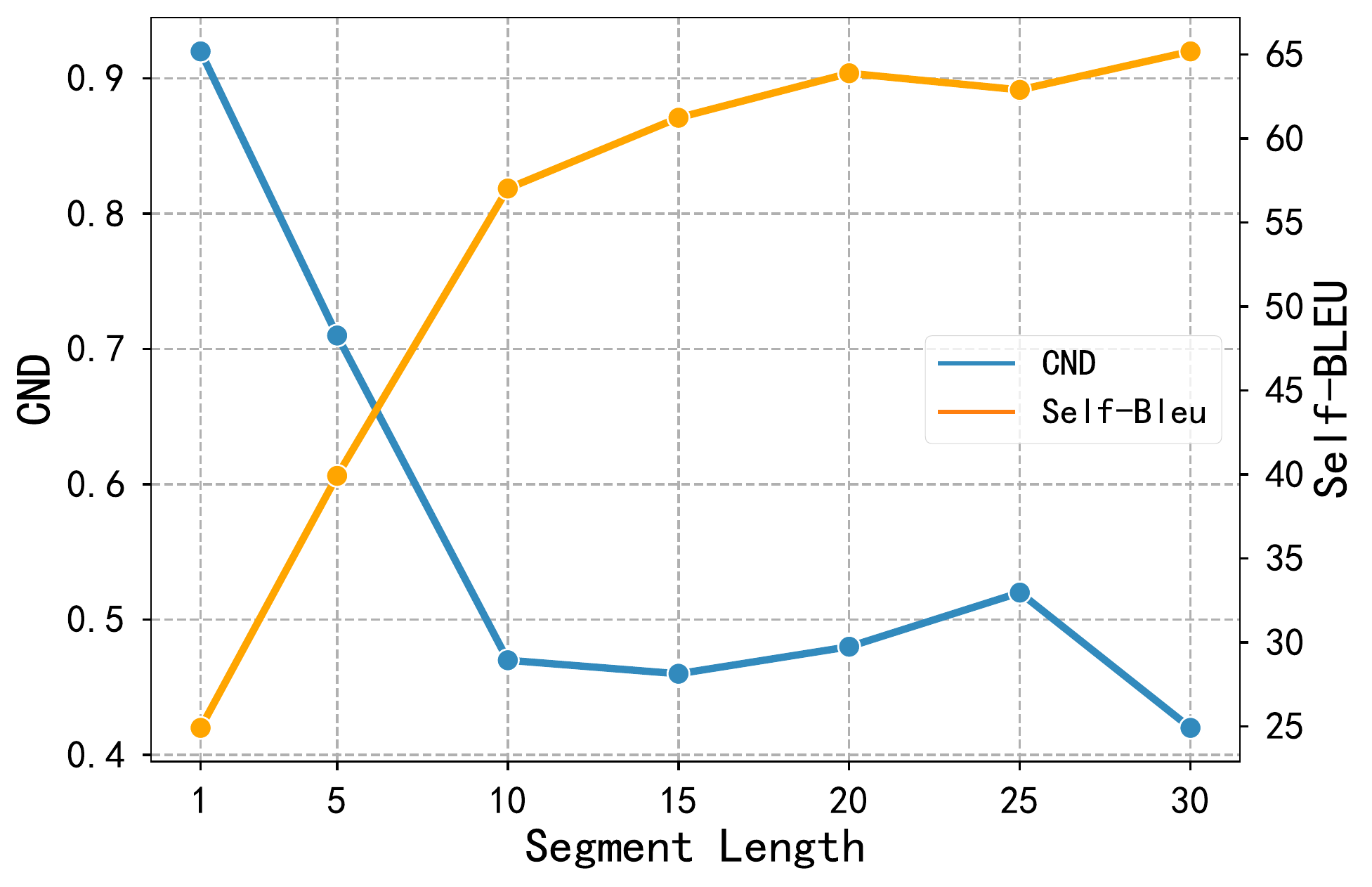}
\caption{CND and Self-Bleu achieved by \modelname-R with varying segment length.}
\label{fig:segment}
\end{minipage}
\end{figure*}

\begin{table}[t]
\centering
\scalebox{0.85}{
\begin{tabular}{c|c|c|c}
\toprule
Model             &    CND$\downarrow$   &    Self-Bleu$\downarrow$   & Dist$\uparrow$   \\ \midrule
\multicolumn{4}{c}{Greedy decoding}   \\ \midrule
IND               &     7.88    &   93.81  &   4.31     \\
CGD               &     55.27   &   99.56  &   0.87     \\ 
No Recurrence     &     4.06    &   94.01  &   4.36     \\
\modelname-R      &     \bf2.01    &   \bf78.11  &   \bf9.28     \\  \midrule
\multicolumn{4}{c}{Beam search, beam size=10}   \\ \midrule
IND               &     76.27   &   97.18  &   0.89    \\
CGD               &     266.52  &   99.82  &   0.08    \\ 
No Recurrence     &     63.12   &   97.41  &   0.85    \\
\modelname-R      &     \bf36.28   &   \bf87.56  &   \bf6.09    \\  \midrule
\multicolumn{4}{c}{Top-k, k=50}   \\ \midrule
IND               &     0.76  &   61.24  &   16.24     \\
CGD               &     0.46  &   68.36  &   13.68     \\ 
No Recurrence    &     \bf0.43  &   65.38  &   15.47     \\
\modelname-R      &     0.47  &   \bf57.02  &   \bf23.01     \\
\bottomrule
\end{tabular}}
\caption{Comparison of different decoding strategies on Yelp Dataset.}
\label{tab:decoding}
\end{table}

\subsection{Analysis}
\label{sec:analysis}
\paragraph{Training Speed} We compare the training speed of \modelname-R and \modelname-P on Yelp dataset with different pre-defined fixed segment lengths. Small segment length will lead to more recurrence steps in \modelname-R. As shown in Fig.~\ref{fig:speed}, small segment length leads to increased number of segments, resulting in much more training time for \modelname-R. In contrast, for \modelname-P, our proposed parallel training method remarkably shortens the training time. When segment length is 1 (token-wise), the training of \modelname-P is more than twice faster than that of \modelname-R. When segment length is 5, \modelname-P is still 50\% faster than \modelname-R. As the segment length grows, the number of segments decreases, and consequently, the training speeds of \modelname-R and \modelname-P reach unanimity. Such empirical results generally confirm the effectiveness of \modelname-P's parallelism. Besides, it is worth emphasizing that despite the deceleration of \modelname-R in the training phase, the inference process is hardly influenced by the recurrence structure because of the auto-regressive decoding manner. Since the efficiency bottleneck of deploying NLG models mainly lies in inference, we believe \modelname-R is still practical enough for downstream NLG tasks.

\paragraph{Separation of Segment} We explore the influence of segment length on the performance of \modelname-R by evaluating CND for generation quality and Self-Bleu for diversity. As Fig. \ref{fig:segment} shows, the generation diversity drops with the increase of segment length. It is consistent with intuition that longer segment leads to looser context correlations but enhanced sampling randomness and thus improved diversity.

\paragraph{Decoding Strategy} We evaluate the performance of \modelname-R on generation quality and diversity with different decoding strategies, including greedy decoding and beam search with beam size $=10$. In these cases, the randomness only originates from sampling the latent variables. As shown in Table \ref{tab:decoding}, by recurrently sampling latent variables which interact with hidden states, \modelname has more distinct advantages in generating diverse text with greedy decoding or beam search. In contrast, sampling decoding itself can enhance generation randomness and thus dilute the diversity improvement obtained by our model. Therefore, using sampling decoding becomes the most difficult case for further improving diversity. We select such a challenging setting in main experiments to verify the effectiveness of \modelname and find it still achieves non-trivial enhancement on diversity, demonstrating the robustness of \modelname in various decoding methods.

\begin{figure}[t]
\centering
\includegraphics[scale=0.46]{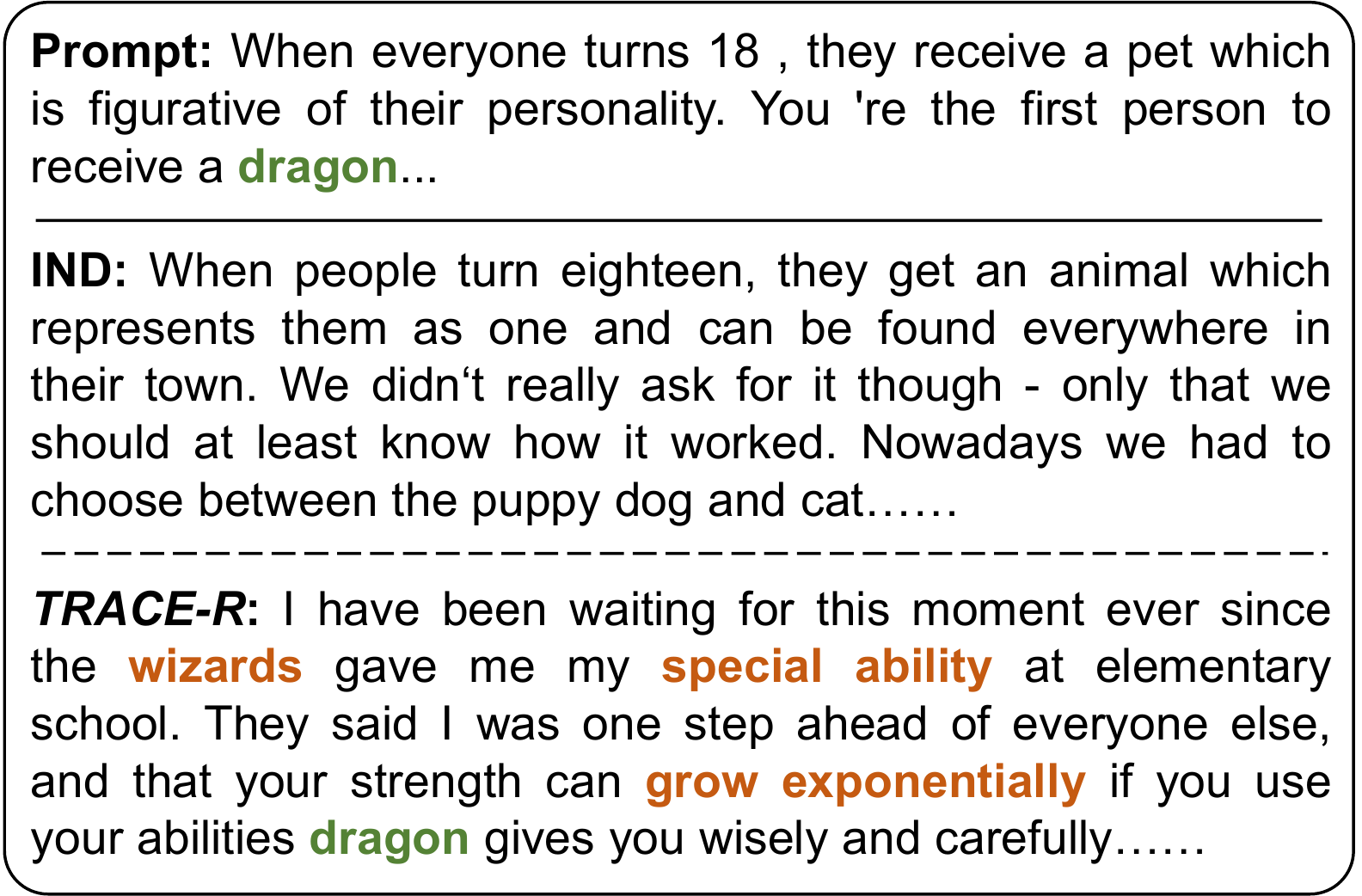} \
\caption{Samples generated by different models on the WP dataset. TRACE-R produces responses to the dragon in the prompt and imagines an engaging story that the protagonist was given a special ability by wizards and would begin an adventure with his pet dragon.}
\label{fig:story_case}
\end{figure}
\subsection{Case Study}
Fig. \ref{fig:story_case} gives one generation example of \modelname-R and IND from WP dataset. The input prompt mentioned that people receive a dragon as a pet. The generated text of IND only talks about receiving an animal. In contrast, the generation of \modelname-R first tells that wizards gave him the ability, and mentions the ability could grow with the dragon. We can see the text produced by our model tells an engaging story like a warrior would go to fight with his given dragon. In general, \modelname-R generates a more vivid continuation compared with IND. 
\section{Conclusion}
In this paper, we revisit the recurrent VAE framework prevalent in the era of RNN, and propose a novel Transformer-based recurrent VAE structure \modelname. \modelname learns a series of segment-wise latent variables conditioned on the preceding ones. We establish the latent distributions with novel residual parameterization. To accelerate training, we design an approximate algorithm of learning latent variables to fit with the Transformer framework. Experimental shows that \modelname achieves significant improvement in generation diversity based on the tight relationship between text hidden states and the latent space. In the future, we will further explore the potential of \modelname in other larger pretrained models like GPT-3.

\section*{Acknowledgement}
Thanks to the anonymous reviewers for their comments. This work is supported by the National Key R\&D Program of China (No. 2020AAA0106502) and Institute Guo Qiang at Tsinghua University.

\section*{Limitations}
While \modelname achieves significant improvement in generation diversity, it still has some limitations. First, the trade-off between quality and diversity is a common problem in natural language generation. \modelname is not an exception. The diversity of \modelname's increases while the quality inevitably drops a little. We will further explore better methods to balance quality and diversity. Second, our parallel acceleration training method requires certain approximations, which, to some extent, hurts the initial advantage of recurrence and leads to a drop in the quality. We will continue to design better acceleration training methods. Third, the speedup of our parallel version of \modelname is limited. Under the practical segment length setting (\textit{e.g.}, 10 and 20), the acceleration is marginal. We also plan to further promote our methods to benefit faster training in the future.

\bibliography{anthology,custom}
\bibliographystyle{acl_natbib}

\newpage
\appendix
\onecolumn
\section{Experiment Details}
\subsection{Metrics Details}
\label{apx_sec:metrics}
\textbf{Perplexity (PPL)}. In the VAE-based framework, we commonly use importance weighted sampling to estimate the PPL. Sampling $n$ latent variables $z_1, z_2, \dots, z_n$ from the variational posterior distribution $q(\boldsymbol{z}_i|\boldsymbol{x})$, we have:
\begin{align}
    \mathcal{L}_{k} =\mathbb{E}\left[\log \frac{1}{n} \sum_{i=1}^{n} \frac{p(\boldsymbol{x},             \boldsymbol{z}_i)}{q(\boldsymbol{z}_i|\boldsymbol{x})}\right]
                    \leq \log \mathbb{E}\left[\frac{1}{n} \sum_{i=1}^{n} \frac{p(\boldsymbol{x}, \boldsymbol{z}_i)}{q(\boldsymbol{z}_i|\boldsymbol{x})}\right]=\log p(\boldsymbol{x}).
\end{align}
As illustrated in \citep{YuriBurda2016ImportanceWA}, when $n\rightarrow\infty$, $\mathcal{L}_n\rightarrow\log p(\boldsymbol{x})$. Therefore, we use $\mathcal{L}_{n}$ to estimate $\log p(\boldsymbol{x})$ and calculate PPL with $n=100$.

\textbf{Mutual Information(MI)} \citep{AlexanderAAlemi2016DeepVI}. Mutual Information $\mathcal{I}(\boldsymbol{x}, \boldsymbol{z})$ measures the mutual dependence between $\boldsymbol{x}$ and $\boldsymbol{z}$, which is defined as:
\begin{align}
&\mathcal{I}_{q}(\boldsymbol{x}, \boldsymbol{z})\!=\!\mathbb{E}_{p(\boldsymbol{x})}\mathbb{E}_{q(\boldsymbol{z}|\boldsymbol{x})} \log q(\boldsymbol{z}|\boldsymbol{x})\!-\!\mathbb{E}_{q(\boldsymbol{z})} \log q(\boldsymbol{z}),
\end{align}
Here, $q_{(\boldsymbol{z})} = \mathbb{E}_{p(\boldsymbol{x})}q(\boldsymbol{z}|\boldsymbol{x})$ is the aggregated posterior.

\textbf{Activate Units(AU)} \citep{YuriBurda2016ImportanceWA}. AU is computed as $A_{\boldsymbol{z}} = \text{Cov}_{\boldsymbol{x}}(\mathbb{E}_{z\sim q(z|\boldsymbol{x})}[z]) > \delta$, where $\delta$ is a threshold. We use AU to measure the active units in latent variables. To more clearly distinguish the performance of \modelname and other baseline models, we increase the threshold to 0.1 from the commonly used 0.01.

\textbf{CND} \citep{JianingLi2020OnTR}. CND measures the similarity between the generated samples and the reference testset by approximating the divergence between these two distributions in n-gram spaces.

\textbf{BLEU} \citep{papineni-etal-2002-bleu}. BLEU evaluates the similarity between the generated text and ground truth by measuring the n-gram overlap of generated samples and references.

\textbf{Rouge} \citep{lin-hovy-2002-manual}. Rouge also calculates the proportion of n-gram overlap between generated examples and reference samples, commonly used in evaluating text summarization. 

\textbf{BERTScore} \citep{TianyiZhang2020BERTScoreET}. BERTScore computes the cosine similarity of representations of generated and reference text obtained by pre-trained BERT \cite{devlin-etal-2019-bert}.

\textbf{Self-BLEU} \citep{YaomingZhu2018TexygenAB}. Self-Bleu computes the BLEU score among the generated text by averaging the BLEU score of each instance with all other samples as references. Lower Self-BLEU means the smaller overlap and higher diversity within the generated samples. 

\textbf{Dist} \citep{li-etal-2016-diversity}. Dist computes the ratio of different n-grams among the generated text.

\textbf{Jaccard Similarity(JS)} \citep{KeWang2018SentiGANGS}. JS computes the Jaccard similarity between every two generated samples.


\subsection{Human Evaluation Details}
\label{sec:hm_detail}
We select 50 prompts from the WP dataset as input to the \modelname and other four baseline models to generate the continuations. We invite five annotators proficient in English to score the generated samples ranging from 1-5 with three criteria: Fluency, Coherence, and Novelty. During the evaluation, each annotator is given 20 groups of samples. Each group has six samples generated by six models, and the order of samples in each group is shuffled to avoid bias. Each sample is scored by two annotators and we average the evaluating score of all samples for each model, reported in the Table \ref{tab:human_eval}.

\section{Additional Proof}
\subsection{Deduction of ELBO of RGD}
\label{proof_rgd}
The lower bound for standard VAE is:
\begin{equation}
    \log p(\boldsymbol{x}) \ge \mathbb{E}_{q(\boldsymbol{z}|\boldsymbol{x})}\log\frac{p(\boldsymbol{x}, \boldsymbol{z})}{q(\boldsymbol{z}|\boldsymbol{x})}
\end{equation}
Rewriting the $q(\boldsymbol{z}|\boldsymbol{x})$ and $p(\boldsymbol{x}, \boldsymbol{z})$ with RGD's factorization, we obtain:
\begin{align}
    \log p(\boldsymbol{x}) \ge& \mathbb{E}_{q(\boldsymbol{z}_{\le T}|\boldsymbol{x}_{\le T})}\log\frac{\prod\limits_{t=1}^{T}p(\boldsymbol{x}_{t}|\boldsymbol{z}_{\le t}, \boldsymbol{x}_{<t})p(\boldsymbol{z}_{t}|\boldsymbol{z}_{<t}, \boldsymbol{x}_{<t})}{\prod\limits_{t=1}^{T}q(\boldsymbol{z}_{t}|\boldsymbol{z}_{<t}, \boldsymbol{x}_{\le t})} \\
    =&\mathbb{E}_{q(\boldsymbol{z}_{\le T}|\boldsymbol{x}_{\le T})}\sum\limits_{t=1}^{T}\log  p(\boldsymbol{x}_{t}|\boldsymbol{z}_{\le t}, \boldsymbol{x}_{<t}) + \log p(\boldsymbol{z}_{t}|\boldsymbol{z}_{<t}, \boldsymbol{x}_{<t}) - \log q(\boldsymbol{z}_{t}|\boldsymbol{z}_{<t}, \boldsymbol{x}_{\le t}) \\
    \label{eq:20}=&\mathbb{E}_{q(\boldsymbol{z}_{\le T}|\boldsymbol{x}_{\le T})}\sum\limits_{t=1}^{T}\log  p(\boldsymbol{x}_{t}|\boldsymbol{z}_{\le t}, \boldsymbol{x}_{<t}) - \log \frac{q(\boldsymbol{z}_{t}|\boldsymbol{z}_{<t}, \boldsymbol{x}_{\le t})}{p(\boldsymbol{z}_{t}|\boldsymbol{z}_{<t}, \boldsymbol{x}_{<t})}
\end{align}
The first term in Eq.\eqref{eq:20} is the construction term of ELBO. For the second term, we have:
\begin{align}
    &\mathbb{E}_{q(\boldsymbol{z}_{\le T}|\boldsymbol{x}_{\le T})}\sum\limits_{t=1}^{T}\log \frac{q(\boldsymbol{z}_{t}|\boldsymbol{z}_{<t}, \boldsymbol{x}_{\le t})}{p(\boldsymbol{z}_{t}|\boldsymbol{z}_{<t}, \boldsymbol{x}_{<t})}\\
    =&\int\prod\limits_{t=1}^{T}q(\boldsymbol{z}_{t}|\boldsymbol{z}_{<t}, \boldsymbol{x}_{\le t})\sum\limits_{t=1}^{T}\log \frac{q(\boldsymbol{z}_{t}|\boldsymbol{z}_{<t}, \boldsymbol{x}_{\le t})}{p(\boldsymbol{z}_{t}|\boldsymbol{z}_{<t}, \boldsymbol{x}_{<t})} \dif \boldsymbol{z}_1\dots\boldsymbol{z}_T \\
    =&\mathbb{E}_{q(\boldsymbol{z}_{\le T-1}|\boldsymbol{x}_{\le T-1})}\mathrm{KL}\Big(q(\boldsymbol{z}_{T}|\boldsymbol{z}_{<T}, \boldsymbol{x}_{\le T})||p(\boldsymbol{z}_{T}|\boldsymbol{z}_{<T}, \boldsymbol{x}_{<T})\Big) + \\
    &\mathbb{E}_{q(\boldsymbol{z}_{\le T-2}|\boldsymbol{x}_{\le T-2})}\mathrm{KL}\Big(q(\boldsymbol{z}_{T-1}|\boldsymbol{z}_{<T-1}, \boldsymbol{x}_{\le T-1})||p(\boldsymbol{z}_{T-1}|\boldsymbol{z}_{<T-1}, \boldsymbol{x}_{<T-1})\Big) +  \\
    &\dots+\mathrm{KL}(q(\boldsymbol{z}_1|\boldsymbol{x}_1)||p(\boldsymbol{z}_1)) \\
    =&\mathbb{E}_{q(\boldsymbol{z}_{\le T}|\boldsymbol{x}_{\le T})}\sum\limits_{t=1}^{T}\mathrm{KL}\Big(q(\boldsymbol{z}_{t}|\boldsymbol{z}_{<t}, \boldsymbol{x}_{\le t})||p(\boldsymbol{z}_{t}|\boldsymbol{z}_{<t}, \boldsymbol{x}_{<t})\Big)
\end{align}
\subsection{Deduction of Parallel Training}
\label{proof1}
\begin{align}
    \boldsymbol{z}_t &= f_{\boldsymbol{\mu}}(\boldsymbol{z}_{t-1}, \boldsymbol{x}_{<t}) + g_{\boldsymbol{\mu}}(\boldsymbol{x}_{t}) + \exp\Big(f_{\boldsymbol{\sigma}}(\boldsymbol{z}_{t-1}, \boldsymbol{x}_{<t}) + g_{\boldsymbol{\sigma}}(\boldsymbol{x}_{t})\Big)\circ \boldsymbol{\epsilon}_t \\
    \label{eq:18}&\approx f_{\boldsymbol{\mu}}(\boldsymbol{z}_{t-1}, \boldsymbol{x}_{<t}) + g_{\boldsymbol{\mu}}(\boldsymbol{x}_{t}) + \Big(f_{\boldsymbol{\sigma}}(\boldsymbol{z}_{t-1}, \boldsymbol{x}_{<t}) + g_{\boldsymbol{\sigma}}(\boldsymbol{x}_{t})+1\Big)\circ \boldsymbol{\epsilon}_t \\
    &= \boldsymbol{W}^f_{\boldsymbol{\mu}1}\boldsymbol{z}_{t-1}+\boldsymbol{W}^f_{\boldsymbol{\mu}2}\boldsymbol{x}_{<t} + \boldsymbol{W}^g_{\boldsymbol{\mu}}\boldsymbol{x}_{t} + \Big(\boldsymbol{W}^f_{\boldsymbol{\sigma}1}\boldsymbol{z}_{t-1}+\boldsymbol{W}^f_{\boldsymbol{\sigma}2}\boldsymbol{x}_{<t} + \boldsymbol{W}^g_{\boldsymbol{\sigma}}\boldsymbol{x}_{t}\Big) \circ \boldsymbol{\epsilon}_t + \boldsymbol{\epsilon}_t \\
    &= \Big[\boldsymbol{W}^f_{\boldsymbol{\mu}2}\boldsymbol{x}_{<t}\!+\! \boldsymbol{W}^g_{\boldsymbol{\mu}}\boldsymbol{x}_{t}\!+\! \Big(\boldsymbol{W}^f_{\boldsymbol{\sigma}2}\boldsymbol{x}_{<t}\!+\! \boldsymbol{W}^g_{\boldsymbol{\sigma}}\boldsymbol{x}_{t}\Big)\!\circ\!\boldsymbol{\epsilon}_t\!+\!\boldsymbol{\epsilon}_t\Big]\!+\!\Big[\boldsymbol{W}^f_{\boldsymbol{\mu}1}\boldsymbol{z}_{t-1}\!+\!\boldsymbol{W}^f_{\boldsymbol{\sigma}1}\boldsymbol{z}_{t-1}\!\circ\!\boldsymbol{\epsilon}_t\Big] \\
    &= \boldsymbol{v}_t + (\boldsymbol{W}^f_{\boldsymbol{\mu}1}+\boldsymbol{W}^f_{\boldsymbol{\sigma}1}\circ\boldsymbol{\epsilon}_t)\boldsymbol{z}_{t-1} \\
    &= \boldsymbol{v}_t + \Big(\boldsymbol{W}^f_{\boldsymbol{\mu}1}+\boldsymbol{W}^f_{\boldsymbol{\sigma}1}\circ\boldsymbol{\epsilon}_t\Big)\Big(\boldsymbol{v}_{t-1} + (\boldsymbol{W}^f_{\boldsymbol{\mu}1}+\boldsymbol{W}^f_{\boldsymbol{\sigma}1}\circ\boldsymbol{\epsilon}_{t-1})\boldsymbol{z}_{t-2}\Big) \\
    &= \boldsymbol{v}_t\!+\!\Big(\boldsymbol{W}^f_{\boldsymbol{\mu}1}\!+\!\boldsymbol{W}^f_{\boldsymbol{\sigma}1}\circ\boldsymbol{\epsilon}_t\Big)\boldsymbol{v}_{t-1}\!+\! \Big(\boldsymbol{W}^f_{\boldsymbol{\mu}1}\!+\!\boldsymbol{W}^f_{\boldsymbol{\sigma}1}\circ\boldsymbol{\epsilon}_t\Big)\Big(\boldsymbol{W}^f_{\boldsymbol{\mu}1}\!+\!\boldsymbol{W}^f_{\boldsymbol{\sigma}1}\circ\boldsymbol{\epsilon}_{t-1}\Big)\boldsymbol{z}_{t-2} \\
    &= \boldsymbol{v}_t\!+\!\sum\limits_{i=1}^{t-1}\prod\limits_{j=i+1}^{t}\Big(\boldsymbol{W}^f_{\boldsymbol{\mu}1}\!+\!\boldsymbol{W}^f_{\boldsymbol{\sigma}1}\circ\boldsymbol{\epsilon}_j\Big)\boldsymbol{v}_{i} \\
    \label{eq:25}&\approx \boldsymbol{v}_t\!+\!\sum\limits_{i=1}^{t-1}\Big((\boldsymbol{W}^f_{\boldsymbol{\mu}1})^n\!+\!(\boldsymbol{W}^f_{\boldsymbol{\sigma}1})^n\circ\boldsymbol{\xi}_j\Big)\boldsymbol{v}_{i} \\
    \label{eq:26}&\approx \boldsymbol{v}_t\!+\!\sum\limits_{i=1}^{t-1}\Big(\boldsymbol{W}^f_{\boldsymbol{\mu}1}\!+\!\boldsymbol{W}^f_{\boldsymbol{\sigma}1}\circ\boldsymbol{\xi}_j\Big)\boldsymbol{v}_{i} = \boldsymbol{v}_t + \sum\limits_{i=1}^{t-1}\boldsymbol{u}_i 
\end{align}
In Eq.\eqref{eq:18}, we use the approximate equality $\exp x \approx x + 1$ when $x$ is close to $0$. In Eq.\eqref{eq:25}, we use one random variable to approximate the multiplication of $t$ independent random variables based on central limit theorem. In Eq.\eqref{eq:26}, we approximate the matrix $\boldsymbol{W}^f_{\boldsymbol{\mu}1}$ and $\boldsymbol{W}^f_{\boldsymbol{\sigma}1}$ as idempotent matrix.

\subsection{Proof of Theorem 1}
\label{proof2}
We expand the KL term as:
\begin{equation}
\begin{aligned}
      &\mathrm{KL}(q(\boldsymbol{z}_{t}|\boldsymbol{z}_{<t}, \boldsymbol{x}_{\le t})||p(\boldsymbol{z}_{t}|\boldsymbol{z}_{<t}, \boldsymbol{x}_{<t})) \\
    =&\frac{1}{2}(\frac{\Delta\boldsymbol{\mu}^2_t}{\boldsymbol{\sigma}^2_t} + \Delta\boldsymbol{\sigma}^2_t - \log \Delta\boldsymbol{\sigma}^2_t -1) 
\end{aligned}
\end{equation}
It is easy to find that $\Delta\boldsymbol{\sigma}^2_t - \log \Delta\boldsymbol{\sigma}^2_t -1 \ge 0$ holds true. So we can control the lower bound of the KL term with $\Delta\boldsymbol{\mu}^2_t$. Following the application of batch normalization to VAE \cite{zhu-etal-2020-batch}, we regulate layer normalization \cite{JimmyBa2016LayerN} to the output of the posterior network, so we have:
\begin{equation}
    \mathbb{E}(\Delta\boldsymbol{\mu}^2_t) \ge \frac{l(\gamma^2 + \beta^2)}{2},
\end{equation}
where $\gamma$ and $\beta$ are the parameters in layer normalization. Therefore, the KL term has a lower bound:
\begin{equation}
    \frac{l(\gamma^2 + \beta^2)}{2\boldsymbol{\sigma}^2_t}.
\end{equation}
\subsection{Proof of Theorem 2}
\label{proof3}
We expand the target upper bound as follow:
\begin{align}
    I_q(\boldsymbol{x}_t;\boldsymbol{z}_t|\boldsymbol{x}_{<t}) =& \iiint q(\boldsymbol{x}_{<t})q(\boldsymbol{x}_t, \boldsymbol{z}_t|\boldsymbol{x}_{<t}) \log\frac{q(\boldsymbol{x}_t, \boldsymbol{z}_t|\boldsymbol{x}_{<t})}{q(\boldsymbol{x}_t|\boldsymbol{x}_{<t})q(\boldsymbol{z}_t|\boldsymbol{x}_{<t})}\dif \boldsymbol{x}_t\dif \boldsymbol{z}_t\dif\boldsymbol{x}_{<t} \\
    =&\iiint q(\boldsymbol{x}_{<t})q(\boldsymbol{x}_t, \boldsymbol{z}_t|\boldsymbol{x}_{<t}) \log\frac{q(\boldsymbol{x}_t|\boldsymbol{x}_{<t}, \boldsymbol{z}_t)} {q(\boldsymbol{x}_t|\boldsymbol{x}_{<t})}\dif \boldsymbol{x}_t\dif \boldsymbol{z}_t\dif\boldsymbol{x}_{<t} \\
    =&\iiint q(\boldsymbol{x}_{<t})q(\boldsymbol{x}_t, \boldsymbol{z}_t|\boldsymbol{x}_{<t}) \log\frac{q(\boldsymbol{x}_t|\boldsymbol{x}_{<t}, \boldsymbol{z}_t)} {p(\boldsymbol{x}_t|\boldsymbol{x}_{<t}, \boldsymbol{z}_t)}\dif \boldsymbol{x}_t\dif \boldsymbol{z}_t\dif\boldsymbol{x}_{<t} + \\
    &\iiint q(\boldsymbol{x}_{<t})q(\boldsymbol{x}_t, \boldsymbol{z}_t|\boldsymbol{x}_{<t}) \log\frac{p(\boldsymbol{x}_t|\boldsymbol{x}_{<t}, \boldsymbol{z}_t)} {q(\boldsymbol{x}_t|\boldsymbol{x}_{<t})}\dif \boldsymbol{x}_t\dif \boldsymbol{z}_t\dif\boldsymbol{x}_{<t} \\
    =&\mathbb{E}_{q(\boldsymbol{z}_t, \boldsymbol{x}_{<t})}\mathrm{KL}(q(\boldsymbol{x}_t|\boldsymbol{x}_{<t}, \boldsymbol{z}_t)||p(\boldsymbol{x}_t|\boldsymbol{x}_{<t}, \boldsymbol{z}_t)) + \\
    &\iiint q(\boldsymbol{x}_{<t})q(\boldsymbol{x}_t, \boldsymbol{z}_t|\boldsymbol{x}_{<t}) \log\frac{p(\boldsymbol{x}_t|\boldsymbol{x}_{<t}, \boldsymbol{z}_t)} {q(\boldsymbol{x}_t|\boldsymbol{x}_{<t})}\dif \boldsymbol{x}_t\dif \boldsymbol{z}_t\dif\boldsymbol{x}_{<t} \\
    \label{eq:33}\ge& \iiint q(\boldsymbol{x}_{<t})q(\boldsymbol{z}_t|\boldsymbol{x}_t, \boldsymbol{x}_{<t})q(\boldsymbol{x}_t|\boldsymbol{x}_{<t}) \log p(\boldsymbol{x}_t|\boldsymbol{x}_{<t}, \boldsymbol{z}_t)\dif \boldsymbol{x}_t\dif \boldsymbol{z}_t\dif\boldsymbol{x}_{<t} - \\
    &\iiint q(\boldsymbol{x}_{<t})q(\boldsymbol{z}_t|\boldsymbol{x}_t, \boldsymbol{x}_{<t})q(\boldsymbol{x}_t|\boldsymbol{x}_{<t}) \log q(\boldsymbol{x}_t|\boldsymbol{x}_{<t})\dif \boldsymbol{x}_t\dif \boldsymbol{z}_t\dif\boldsymbol{x}_{<t} \\
    \label{eq:35}\ge& \iiint q(\boldsymbol{x}_{<t})p(\boldsymbol{z}_t|\boldsymbol{x}_t, \boldsymbol{x}_{<t})q(\boldsymbol{x}_t|\boldsymbol{x}_{<t}) \log p(\boldsymbol{x}_t|\boldsymbol{x}_{<t}, \boldsymbol{z}_t)\dif \boldsymbol{x}_t\dif \boldsymbol{z}_t\dif\boldsymbol{x}_{<t} \\
    =& \mathbb{E}_{q(\boldsymbol{z}_{\le t}|\boldsymbol{x}_{\le t})}\log p(\boldsymbol{x}_t|\boldsymbol{z}_{\le t}, \boldsymbol{z}_t).
\end{align}
In Eq.\eqref{eq:33}, the inequality holds true because the KL value is always great than $0$. In Eq.\eqref{eq:35}, the inequality holds true because $-\log q(\boldsymbol{x}_t|\boldsymbol{x}_{<t})$ is always greater than $0$.

\end{document}